\documentclass[runningheads]{llncs}
\usepackage[utf8]{inputenc}
\usepackage{array}
\usepackage{float}
\usepackage{amstext}
\usepackage{graphicx}
\PassOptionsToPackage{normalem}{ulem}
\usepackage{ulem}
\usepackage[bookmarks=false,
 breaklinks=false,pdfborder={0 0 1},backref=page,colorlinks=false]
 {hyperref}
\hypersetup{
 pagebackref,breaklinks,colorlinks,citecolor=eccvblue}

\makeatletter

\DeclareTextSymbolDefault{\textquotedbl}{T1}
\providecommand{\tabularnewline}{\\}
\floatstyle{ruled}
\newfloat{algorithm}{tbp}{loa}
\providecommand{\algorithmname}{Algorithm}
\floatname{algorithm}{\protect\algorithmname}

\usepackage{eccv}

\usepackage{eccvabbrv}

\usepackage[accsupp]{axessibility}

\usepackage{orcidlink}

\usepackage[linesnumbered,algo2e]{algorithm2e}
\newcommand{\hrulealg}[0]{\vspace{1mm} \hrule \vspace{1mm}}
\author{Tuyen Tran}

\authorrunning{Tuyen Tran}
\titlerunning{Spatial-temporal Refinement for Consistent Semantic Segmentation}

\institute{Applied Artificial Intelligence Institute, Deakin University, Australia\\Team: TXT\\
\email{t.tran@deakin.edu.au}}

\makeatother

\begin{document}

\title{The 2nd Solution for LSVOS Challenge RVOS Track: Spatial-temporal
Refinement for Consistent Semantic Segmentation}
\maketitle
\begin{abstract}
Referring Video Object Segmentation (RVOS) is a challenging task due
to its requirement for temporal understanding. Due to the obstacle
of computational complexity, many state-of-the-art models are trained
on short time intervals. During testing, while these models can effectively
process information over short time steps, they struggle to maintain
consistent perception over prolonged time sequences, leading to inconsistencies
in the resulting semantic segmentation masks. To address this challenge,
we take a step further in this work by leveraging the tracking capabilities
of the newly introduced Segment Anything Model version 2 (SAM-v2)
to enhance the temporal consistency of the referring object segmentation
model. Our method achieved a score of $60.40$ $\mathcal{J\text{\&}F}$
on the test set of the MeViS dataset, placing 2nd place in the final
ranking of the RVOS Track at the ECCV 2024 LSVOS Challenge.
\end{abstract}

\section{Introduction \protect\label{sec:intro}}

This technical report details our participation in the 6th LSVOS challenge,
which offers two tracks: the VOS (video object segmentation) track
and the RVOS (referring video object segmentation) track. This year's
challenge has introduced new datasets, MOSE \cite{MOSE} and MeViS
\cite{MeViS}, replacing the YouTube-VOS and Referring YouTube-VOS
benchmarks used in the past. These new datasets bring more complex
scenarios, including disappearing and reappearing objects, small and
difficult-to-detect objects, significant occlusions, crowded settings,
and longer video durations, making the competition tougher than previous
years.

In this work, we focus on the RVOS task. RVOS involves locating target
instances throughout an entire video sequence based on a linguistic
descriptor. The main challenge of this task, compared to image-based
segmentation, is that it requires the machine learning model to understand
the temporal relationships within the video. The recently introduced
Motion Expression guided Video Segmentation (MeViS) dataset \cite{MeViS}
places additional emphasis on the importance of machine learning model's
capability in temporal understanding. Compared to earlier datasets,
MeViS is more challenging, as most language expressions in the dataset
incorporate motion elements. Motion can occur within short time intervals
or over the entire course of a video, requiring the machine learning
model to comprehensively understand multi-modal interactions throughout
long-term sequences.

Aware of this challenge, recent works have focused on enhancing the
temporal understanding of machine learning models. The latest state-of-the-art
in RVOS, DsHmp \cite{DsHmp}, addresses this issue by separating the
referring description into static and motion cues. They then employ
a hierarchical motion perception module to interpret the visual information.
The long visual tokens are processed hierarchically, enabling improved
temporal comprehension. In the RVOS Track of the CVPR 2024 PVUW Challenge
\cite{ding2024pvuw}, the TIME Team \cite{Pan20243rdPS} proposed
a novel approach for referring video object segmentation. They first
use a frozen CLIP \cite{radford2021learning} model for video and
language feature extraction. Then, they introduce a query initialization
method to generate high quality queries for both frames and videos.
Taking a different approach, the Tapall.ai team \cite{gao20241st}
fine-tuned the current leading RVOS model, MUTR \cite{yan2024referred},
on the MeViS dataset. During inference, they explored various frame
sampling strategies to enhance the baseline method. The CASIA IVA
team \cite{Cao20242ndPS} also used MUTR as the base model to extract
the initial coarse masks. They then enhanced the consistency of the
predicted results by introducing proposal instance masks into the
model for query initialization. In the final stage, they utilized
the segmentation capabilities of SAM-HQ \cite{sam_hq} for spatial
refinement.

In this work, we also focus on improving the consistency of segmentation
results from the base model. We use SAM-v2 \cite{ravi2024sam2} to
refine the initial coarse segmentation, not just in the spatial dimension.
Instead, we leverage the advanced segmentation and tracking capabilities
of SAM-v2 for both spatial and temporal refinement. SAM-v2 is a newly
introduced model. In addition to its strong segmentation capabilities,
SAM-v2 has demonstrated excellent performance in tracking instances
over long videos. We utilize the spatio-temporal masks (i.e., 'masklets')
generated by SAM-v2 to enhance the consistency of segmentation masks
across both spatial and temporal dimensions. Our solution achieved
$53.66$ $\mathcal{J\text{\&}F}$ on the validation set and $60.40$
$\mathcal{J\text{\&}F}$ on the test set of the MeViS dataset. 

\section{Method}

\begin{figure}
\begin{centering}
\includegraphics[width=0.99\textwidth]{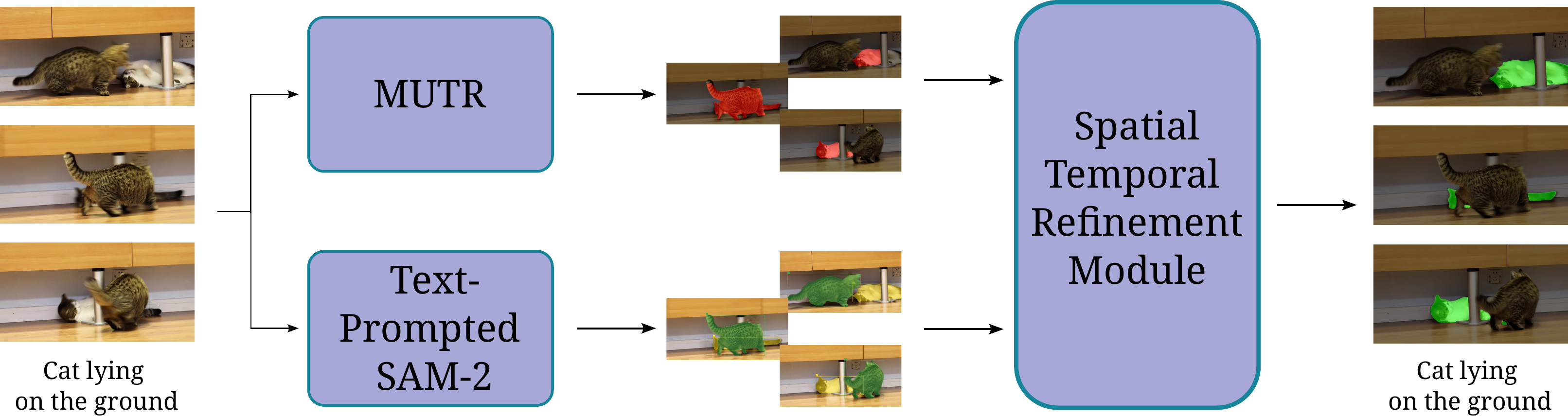}
\par\end{centering}
\caption{We first extract the main noun from the given textual query (e.g.,
\textquotedbl Cat\textquotedbl ) and use it as input for the \emph{Text-Prompted
SAM-2}. This module essentially combines Grounding Dino and SAMv2.
Grounding Dino detects all bounding boxes of instances belonging to
the specified object category. These boxes are then used as prompt
input for the SAMv2 model, resulting in a sequence of spatio-temporal
masks. Concurrently, a fine-tuned MUTR model is employed to generate
coarse masks from the input video. These initial masks are then subjected
to refinement within the Spatial-Temporal Refinement Module, resulting
in final segmentation masks with improved temporal consistency.\protect\label{fig:overal}}
\end{figure}

The overall proposed framework is presented in Figure \ref{fig:overal}.
We initially employ SAMv2 to extract spatio-temporal masks containing
tracking-related details. Simultaneously, we fine-tune the MUTR model
on MeViS to generate initial coarse spatio-temporal masks based on
the given video and textual description. We define these coarse masks
as $M_{c}=\left\{ u^{t}\right\} _{t=1}^{T},$ where $T$ is the number
of frame in video. The resulting raw masks undergo further refinement
in the Spatial-Temporal Refinement Module (Section 2.2) to yield the
final segmentation mask with enhanced temporal consistency.

\subsection{Video object tracking with textual prompt}

Since the original SAM-2 requires initial inputs of either points,
boxes, or masks to track visual objects within a video, additional
processing steps are necessary to construct a video object tracking
system using SAM-2 with textual prompt input. First, given a descriptive
sentence, we employ a language processing tool Berkeley Neural Parser
\cite{kitaev} to extract the main noun (e.g., \textquotedbl Cat\textquotedbl ,
in Figure \ref{fig:overal}), which designates the target object category
for tracking. Subsequently, we utilize the open-vocabulary object
detection model, Grounding DINO \cite{liu2023grounding}, to extract
bounding boxes encompassing the target object. These bounding boxes
serve as input prompts for the SAM-2 model. SAM-2 produces a set of
spatio-temporal masks, termed 'masklets'. The number of masklets corresponds
to the quantity of distinct instances detected for the given object
category. Formally, we denote the tracking results from SAM-2 as $M_{t}=\left\{ v_{i}^{t}\right\} $,
where i ranges from $1$ to $N$ and $t$ ranges from $1$ to $T$.
Here, $N$ represents the number of instances detected for the given
object category, and $T$ denotes the number of frames in the input
video.

\subsection{Spatial-Temporal Refinement for Consistent Semantic Segmentation}

\begin{figure}
\begin{centering}
\includegraphics[width=0.99\textwidth]{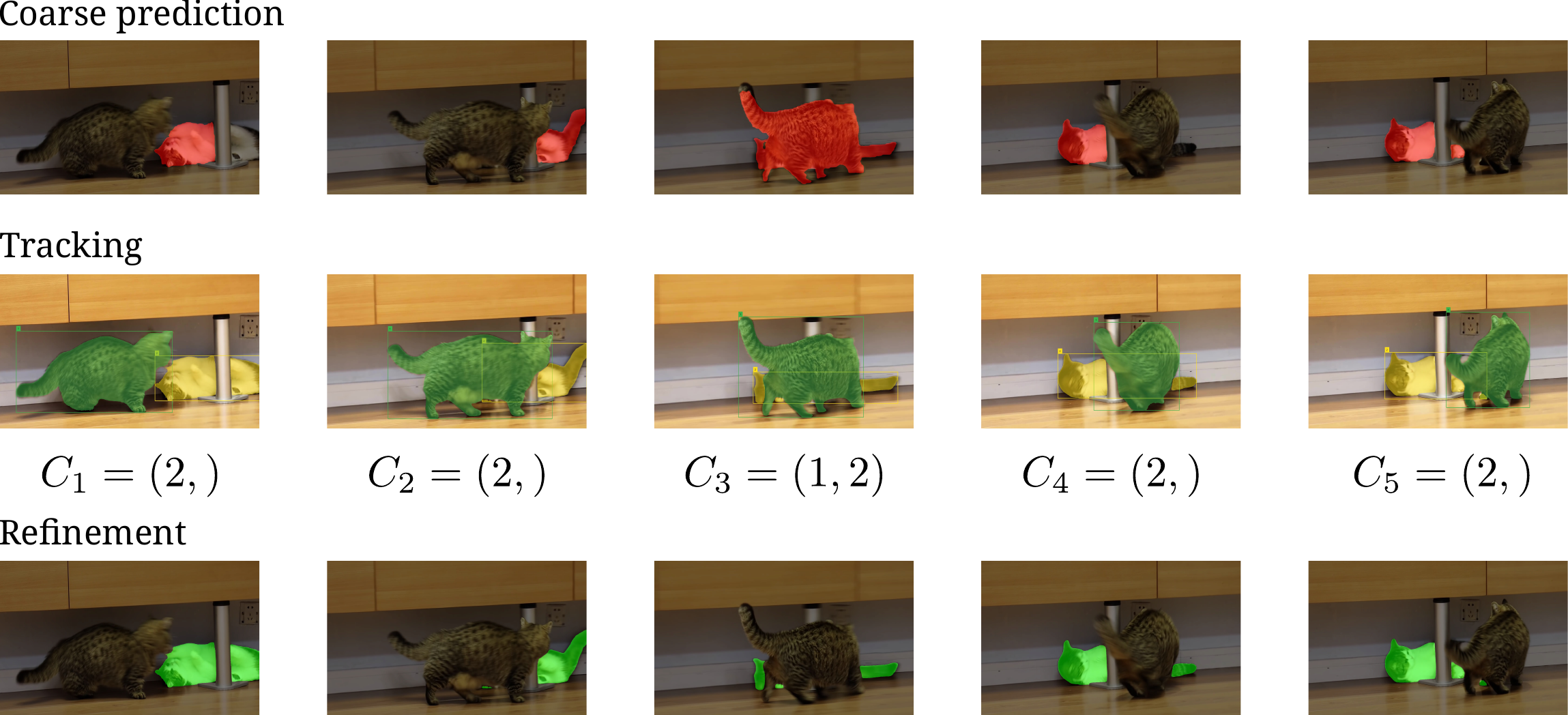}
\par\end{centering}
\caption{Spatial-temporal Refinement Algorithm: We utilize the coarse prediction
masks $\left\{ u^{t}\right\} $ and tracked masks $\left\{ v_{i}^{t}\right\} $
to construct component combinations $C_{t}$ for each time step $t$.
In the example shown, during time steps $1,2,4$, and $5,$ the coarse
prediction from baseline consistently segments only the cat tracked
with ID $2$ (yellow mask). However, at time step $3$, both cats
are segmented, resulting in $C_{3}=(1,2)$. To maintain the temporal
consistency, we select the most frequent combination, $(2,)$ and
apply it to refine all frames within the window size. \protect\label{fig:method}}
\end{figure}

The pseudo code is outlined in Algorithm \ref{alg:psedu-code}. We
first divide the entire video, which consists of $T$ frames, into
non-overlapping sequences with a window size of $W$. The proposed
module, which takes input as coarse masks $M_{c}$ and tracked masks
$M_{t}$, is executed on each sequence individually. With slight abuse
of notation, we also will use $M_{t}=\left\{ v_{i}^{t}\right\} $
with $i$ ranges from $1$ to $W$ to denote the tracking results
of instance $v_{i}$ within the window size $W$. At each time step,
we calculate the Fraction of Overlap $f_{i}^{t}$ between each tracked
instance $v_{i}^{t}$ and the coarse segmentation mask $u^{t}$:

\begin{equation}
f_{i}^{t}=\frac{\text{Intersection\ensuremath{\left(v_{i}^{t},u^{t}\right)}}}{\text{Area(\ensuremath{v_{i}^{t}})}}.
\end{equation}
$f_{i}^{t}$ is calculated as the ratio of the intersection area between
the tracked instance $v_{i}^{t}$ and the coarse mask $u^{t}$ to
the total area of instance $v_{i}^{t}$. The Fraction of Overlap $f_{i}^{t}$
indicates the proportion of instance $i$ at time step $t$ that overlaps
with the coarse mask predicted by the MUTR model. If $f_{i}^{t}$
exceeds a threshold $\tau$, we infer that instance $i$ is present
at time step $t$ and add its index to the component list $C_{t}$.
As a result, $C_{t}$ represents the combination of components at
the time step t. This process is repeated across all time steps within
the window size $W$, yielding the set $\mathcal{C}=\left\{ C_{t}\right\} ,$
where each element $C_{t}$ captures the specific combination of components
at its respective time step $t$. We expect that within a given window
size, the spatio-temporal masks should remain consistent. Therefore,
we select $C_{\text{sel}}$ as the combination of components that
appears most frequently in the set $\mathcal{C}$ for refinement.
The refined spatio-temporal masks $M_{r}=\left\{ m^{t}\right\} $
are derived by composing all instances listed in $C_{\text{sel}}$.
If $C_{\text{sel}}=\emptyset$, meaning that the predicted instances
from the MUTR model are not included in the tracking output, we retain
the original prediction without refinement. We illustrate the behavior
of the proposed refinement algorithm with an example in Figure \ref{fig:method}.

\SetKwComment{Comment}{/* }{ */} 

\begin{algorithm*}
\KwIn{
\begin{itemize}
\item[-] Coarse segmentation masks $M_{c}=\left\{ u^{t}\right\} _{t=1}^{W},$
\item[-] Tracked masks $M_{t}=\left\{ v_{i}^{t}\right\} _{i=1,t=1}^{N,W}$
\end{itemize}
}

\KwOut{
\begin{itemize}
\item[-] Refined mask $M_{r}=\left\{ m^{t}\right\} _{t=1}^{W}$ 
\end{itemize}
}

\hrulealg

\For{$t=1$ to $W$} {

\ForEach{$v_{i}^{t}$}{

Calculate $f_{i}^{t}$ \tcp*[f]{Refer to equation 1}\;

\lIf{$f_{i}^{t}>\tau$}{Add $i$ to $C_{t}$ }

} }

Set $C_{\text{sel}}$ as the combination that appears most frequently
in $\left\{ C_{t}\right\} $\; 

Obtain the refined mask $M_{r}=\left\{ m^{t}\right\} _{t=1}^{W}$
by composing all instances included in $C_{\text{sel}}$\;

\caption{Spatial-Temporal Refinement for Consistent Segmentation.\protect\label{alg:psedu-code}}
\end{algorithm*}

\section{Experiment}

\subsection{Implementation details}

We used open-source models from SAM-2 and Grounding DINO for tracking
objects in video based on textual prompts. For fine-tuning MUTR, we
followed the approach in \cite{gao20241st}, utilizing MUTR’s weights
that were trained on both the image \cite{kazemzadeh2014referitgame},
and video \cite{seo2020urvos} referring datasets. We fine-tuned MUTR
for three epochs using four NVIDIA A100 GPUs, each with 40GB of VRAM.
When refining the initial coarse masks, we selected a window size
$W$ of $15$ and a threshold $\tau$ of $0.8$.

\subsection{Quantitative results}

\subsubsection*{Main results:}

The main results is present in Table \ref{tab:main_result}. Our method
achieve $60.40$ $\mathcal{J\text{\&}F}$ in the test set of RVOS
Track at the ECCV 2024 LSVOS Challenge. placing the 2nd position in
the leader board 

\begin{table}
\begin{centering}
\begin{tabular}{>{\raggedleft}m{0.08\textwidth}>{\raggedleft}m{0.32\textwidth}|>{\centering}m{0.16\textwidth}>{\centering}m{0.16\textwidth}>{\centering}m{0.16\textwidth}}
\hline 
Place & Team\hspace*{0.6cm} & $\mathcal{J\text{\&}F}$ & $\mathcal{J}$ & $\mathcal{F}$\tabularnewline
\hline 
\hline 
1 & times\hspace*{0.4cm} & $62.57$ & $58.98$ & $66.15$\tabularnewline
2 & \uline{tuyentx}\hspace*{0.4cm} & $\underline{60.40}$ & $\underline{60.40}$ & $\underline{63.78}$\tabularnewline
3 & Jasonshelter0\hspace*{0.4cm} & $60.36$ & $56.88$ & $63.85$\tabularnewline
4 & SaBoTaGe\hspace*{0.4cm} & $60.36$ & $56.89$ & $63.83$\tabularnewline
5 & BBBiiinnn\hspace*{0.4cm} & $60.36$ & $56.88$ & $63.84$\tabularnewline
6 & bdc\hspace*{0.4cm} & $60.28$ & $56.82$ & $63.74$\tabularnewline
\hline 
\end{tabular}
\par\end{centering}
\vspace{1mm}

\caption{Quantitative results on the MeViS test set.\protect\label{tab:main_result}}
\end{table}

\subsubsection*{Ablation study:}

We conducted an ablation study on the MeViS validation set in Table
\ref{tab:ablation}. Using the coarse masks from MUTR resulted in
a score of $48.38$$\mathcal{J\text{\&}F}$ on the validation set.
Incorporating tracked masks obtained from SAM-2 notably improved performance,
with a window size of $15$ yielding the best results.

\begin{table}
\begin{centering}
\begin{tabular}{>{\raggedleft}m{0.16\textwidth}|>{\centering}m{0.16\textwidth}>{\centering}m{0.16\textwidth}>{\centering}m{0.16\textwidth}>{\centering}m{0.16\textwidth}>{\centering}m{0.16\textwidth}}
\hline 
Method\hspace*{0.6cm} & Baseline & $W=5$ & \textbf{$W=10$} & $W=15$ & $W=20$\tabularnewline
\hline 
\hline 
$\mathcal{J\text{\&}F}$\hspace*{0.6cm} & $48.38$ & $50.16$ & $52.20$ & $\underline{53.66}$ & $51.33$\tabularnewline
\hline 
\end{tabular}
\par\end{centering}
\vspace{3mm}

\caption{Ablations on MeViS validation set.\protect\label{tab:ablation}}
\end{table}

\subsection{Qualitative results}

\begin{figure}
\begin{centering}
\includegraphics[width=0.99\textwidth]{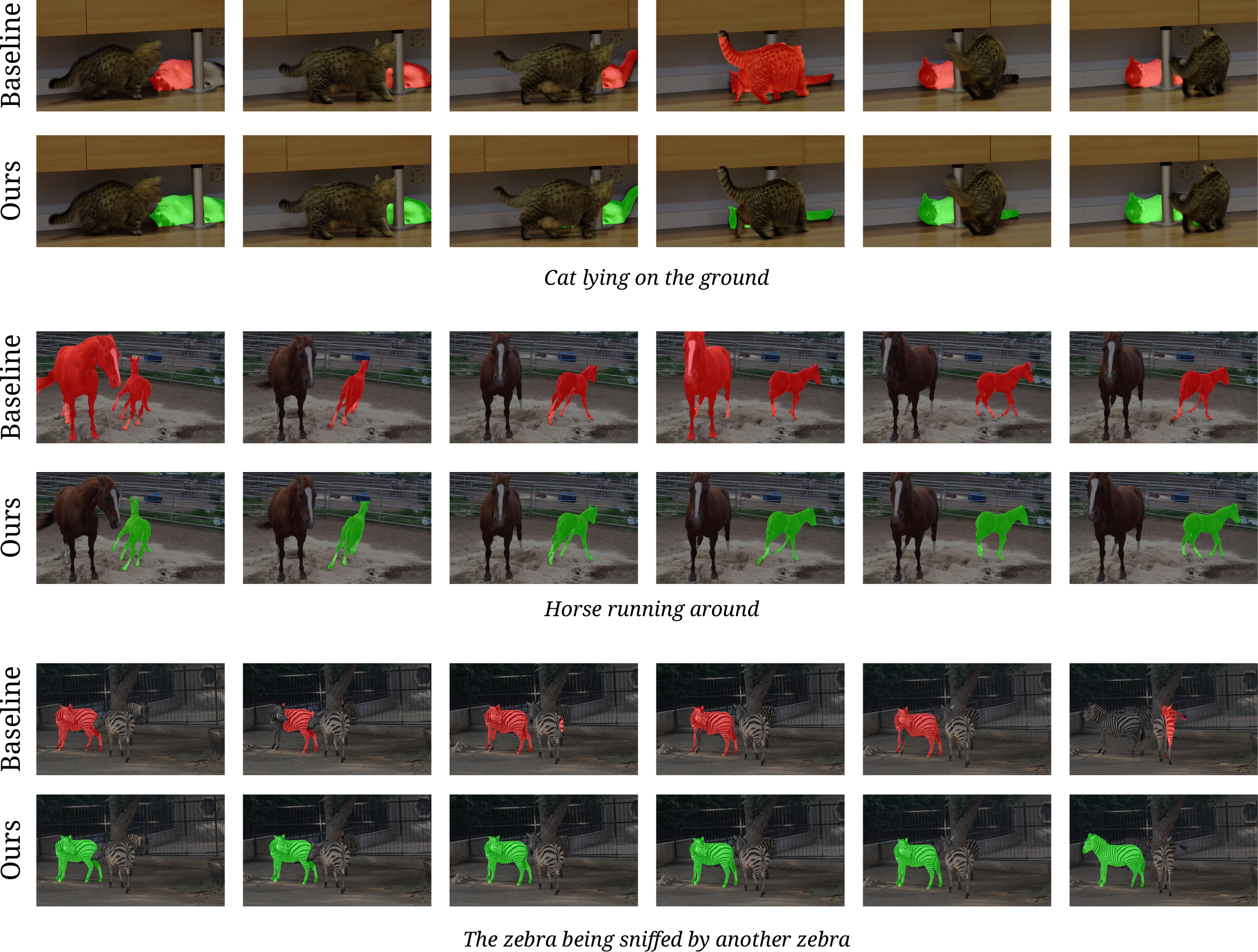}
\par\end{centering}
\caption{Qualitative results on the MeViS validation set: We present examples
to showcase the effectiveness of the proposed approach. The baseline
is initial coarse masks obtained from MUTR. It is shown that while
the baseline can produce accurate masks for short periods of time,
it struggles to maintain consistency over longer periods. By utilizing
the tracking capabilities of SAM-2, we refine the initial coarse masks
to achieve improved consistency across both spatial and temporal dimensions.
\protect\label{fig:qualitative}}
\end{figure}

We showcase qualitative results in Figure \ref{fig:qualitative},
which illustrates the ability of the proposed method to enhance the
temporal consistency of semantic segmentation compared to the baseline
model.

\section{Conclusion}

In this work, we present a simple yet effective method for segmenting
instances in video based on a referring description. We build upon
the baseline by leveraging the tracking capabilities of the SAM-2
model to enhance the temporal consistency of the segmentation results.
By incorporating tracking information from an external model, we partially
address the limitations in long-term temporal understanding seen in
current state-of-the-art methods for this challenging task. Although
the improved performance over baselines in both quantitative and qualitative
evaluations demonstrates the effectiveness of our approach, it also
highlights the need for end-to-end models with enhanced long-term
understanding.

\bibliographystyle{splncs04}
\bibliography{main}
 
\end{document}